\documentclass[sigconf, nonacm]{acmart}

\usepackage{booktabs} 
\usepackage{graphicx}
\usepackage{subcaption}
\usepackage{xcolor,soul}
\usepackage[utf8]{inputenc}

\usepackage{multirow}
\usepackage{csvsimple}

\begin{document}
\twocolumn
\title{Mobile Recognition of Wikipedia Featured Sites using Deep Learning and Crowd-sourced Imagery}

\author{Jimin Tan}
\affiliation{New York University}
\email{tanjimin@nyu.edu}

\author{Anastasios Noulas}
\affiliation{New York University}
\email{noulas@nyu.edu}

\author{Diego Sáez}
\affiliation{Wikimedia Foundation}
\email{diego@wikimedia.org}

\author{Rossano Schifanella}
\affiliation{University of Turin}
\email{schifane@di.unito.it}

\begin{abstract}
Rendering Wikipedia content through mobile and augmented reality mediums can enable new forms of interaction in urban-focused user communities
facilitating learning, communication and knowledge exchange. With this objective in mind, in this work we develop a mobile application that allows for the recognition of notable sites featured on Wikipedia.  The application is powered by a deep neural network that has been trained on crowd-sourced imagery describing sites of interest, such as buildings, statues, museums or other physical entities that are present and visually accessible in an urban environment. We describe an end-to-end pipeline that describes data collection, model training and evaluation of our application considering online and real world scenarios. We identify a number of challenges in the site recognition task which arise due to visual similarities amongst the classified sites as well as due to noise introduce by the surrounding built environment. We demonstrate how using mobile contextual information, such as user location, orientation and attention patterns can significantly alleviate such challenges. Moreover, we present an unsupervised learning technique to de-noise crowd-sourced imagery which improves classification performance further.

\end{abstract}

\maketitle

\section{Introduction}

The capability to execute computationally demanding tasks on mobile devices thanks to new software and hardware technology has paved the way for the emergence of a new generation of user facing applications that are powered by artificial intelligence modules and augmented reality features. Applications that have been enabled through this paradigm include mobile and wearable systems that recognise objects to trigger events and facilitate exploration of the user's environment, infer social dynamics or track the psychological patterns of individuals~\cite{spathis2019sequence, seneviratne2017survey, georgiev2014dsp}. Object recognition using mobile devices in particular has become rather popular due to the multitude of applications it can support~\cite{googleLens, van2018inaturalist}. This new generation of apps can enhance awareness of user surroundings and context, leading to a better experience in tasks including search, navigation and transport, entertainment, translation and more. 

In this work, we are focusing on the problem of identifying notable sites, including landmarks or other visually recognisable points of interest, at large geographic scales. While the problem of site recognition has been explored in small geographic regions e.g. to support touristic experiences in old towns or archeological sites~\cite{neuburger2017afternoon,billinghurst2015survey,dahne2002archeoguide}, extending it to larger areas has been a challenge. First and foremost, the availability of data sets that consistently provide high quality imagery for model training and associate that visual information with the geographic location of outdoor sites has been scarce. Furthermore, in dense areas, where multiple target sites can be present, and in proximity to one another, the task of distinguishing a unique object can become challenging. The effect becomes more severe as sites can be similar (e.g. two statues in the same square), or the fact that non relevant objects can trigger false positives which may harm user experience. Finally, deploying an object identification task outdoors involves difficulties that have to do with varying weather conditions or rendering imagery during nighttime. These environmental conditions effectively translate to a noisy environment that the system in place has to cope with~\cite{zhou2014learning, kendall2016modelling}. 

To obtain a list of notable sites we employ Wikipedia~\cite{Wikipedia} as our primary data source. Wikipedia is a crowdsourced knowledge base that grows continuously, explictly focusing on subjects ranging from general to a more specialized interest. 
We can therefore exploit it to extract a list of relevant sites and to provide contextual knowledge to enrich the exploration of a landmark or a point of interest in a city.
Secondly, visual information describing notable sites is easier to access. It is thus possible to obtain high quality training data that we use as input to train deep learning models to effectively recognise a site.  Finally, a significant fraction of Wikipedia articles referring to sites that lie in some geography are geo-tagged ($\sim 30\%$ of all English Wikipedia articles are geo-tagged~\cite{wikiDumps}). As a result incorporating Wikipedia content on mobile applications that enable exploration in geographic space becomes easier, in addition to being able to utilize contextual geographic information about the user to enhance the visual recognition task as we show next.
In summary, we make the following key contributions:
\begin{itemize}
	\item \textbf{Designing an end-to-end Mobile System for Outdoor Site Recognition:} From data collection to mobile development and deployment, 
	we design a system pipeline to power a mobile application that enables the exploration and recognition of historic and interesting sites in urban environments. 
	We use a Convolution Neural Network (CNN) architecture~\cite{lecun1995convolutional} to train deep site recognition models which are specialised at identifying sites at particular geographic regions. When a user navigates to a particular area, the corresponding region-specific model is off-loaded to the user's device.
	To allow for swift execution in a mobile setting we choose a minimal model architecture~\cite{iandola2016squeezenet} that maintains high prediction accuracy and at the same time keeps the number of training parameters small. This results to a lite model with small memory and network bandwidth footprint, appropriate for mobile deployments and when users are moving.
	\item \textbf{Filtered Training on Crowd-sourced Images:} We train deep learning models for site recognition on images obtained through two different sources, Google Images~\cite{googleImages} and Flickr~\cite{flickr}. In reflection to the observation that crowd-sourced image data can vary in quality or be irrelevant for the intended use in the present work, we introduce an unsupervised image de-noising method that improves prediction accuracy by a significant margin when compared to a baseline trained on the raw set of collected imagery.
	The method applies Jensen-Shannon Divergence~\cite{lin1991divergence} on features extracted by a CNN model pre-trained on ImageNet to detect outlier images and remove them from the training set. Following this approach almost doubles prediction accuracy in the site recognition task. We then evaluate our system across a variety of verticals important with regard to the application scenario envisioned. Initially we consider an \textit{online} evaluation scenario where training, testing and validation takes places considering items from a given input dataset. We examine variations in terms of the difficulty of the site identification task in terms of number of candidate items in the prediction space, the typology of the sites considered (buildings, statues museums etc.) as well as their respective popularity.
	\item \textbf{Incorporating Mobile Context Awareness:} Through a test deployment of our mobile application we observe a large discrepancy in prediction accuracy terms between testing on the data collected from online sources and input image items provided by mobile users captured in realistic navigation experiences near sites of interest. We perform an interpretability analysis on the recognition tasks using the Grad-CAM system proposed in~\cite{selvaraju2017grad} observing that built structures surrounding the site of interest to the user can \textit{confuse} the classifier. In reflection of these observations, to assist the visual classifier on the site recognition task, we incorporate mobile contextual information. We show how combining user's current location and orientation information with a simple focus interaction on the rendered site of interest, prediction accuracy can dramatically improve allowing for the correct recognition of two in every three sites tested.
\end{itemize}
Applications such as the one we built in the present work can power a new generation of mobile services that can enable new forms of urban exploration and facilitate accessibility to local knowledge to tourists and city residents alike. Furthermore, rendering Wikipedia content using mobile as medium 
can enhance the exploration of encyclopedic content enabling novel forms of interaction. with physical space.
This approach also improves over the limitations of text-based search and web browsing that can be frustrating when moving, disconnecting users from their surroundings. Next in Section~\ref{systemarch} we provide an overview of our system. In Section~\ref{onlineexperiments} we describe the training methodology we adhere to using crowdsourced data and a filtering method to de-noise input imagery. In Section~\ref{sec:mcontext} we run a test-bed of our system in the wild and introduce a mobile-context awareness mechanism for the site recognition task and in Section~\ref{limitations} we discuss the limitations of and future vision for our system.
Finally, we discuss related work in Section~\ref{related} and conclude our work in Section~\ref{conclusion}.

\section{Mobile Site Recognition}
\label{systemarch}
In this section we provide an overview of the architecture of a site recognition system embedded in a mobile application we develop together with a description of a crowdsourced image data collection pipeline that is critical for training such system.
\begin{figure}[t]
\centering
\includegraphics[scale=0.2] {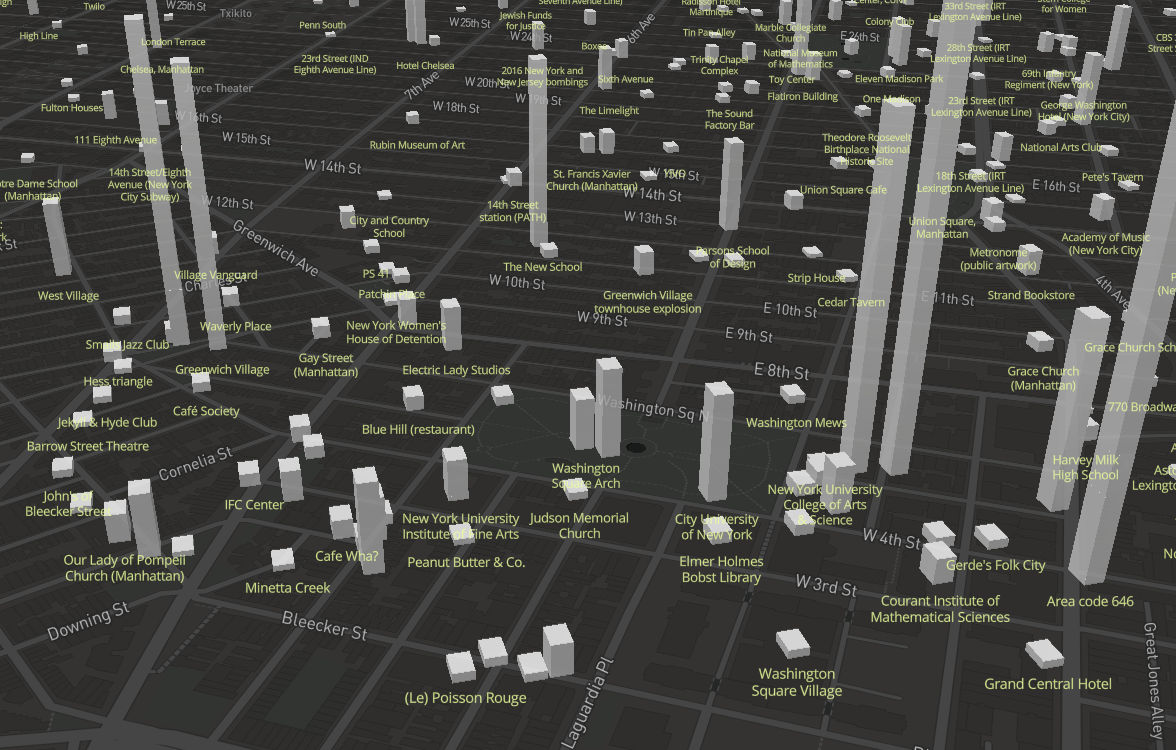}
\caption{A set of sites featured on Wikipedia in a neighborhood of New York City. Each site is represented by a 3-d rectangle with its height corresponding to the site's pageview count on Wikipedia.}
\label{wikimapnyc}
\end{figure}

\begin{figure}[t]
\centering
\includegraphics[scale=0.15] {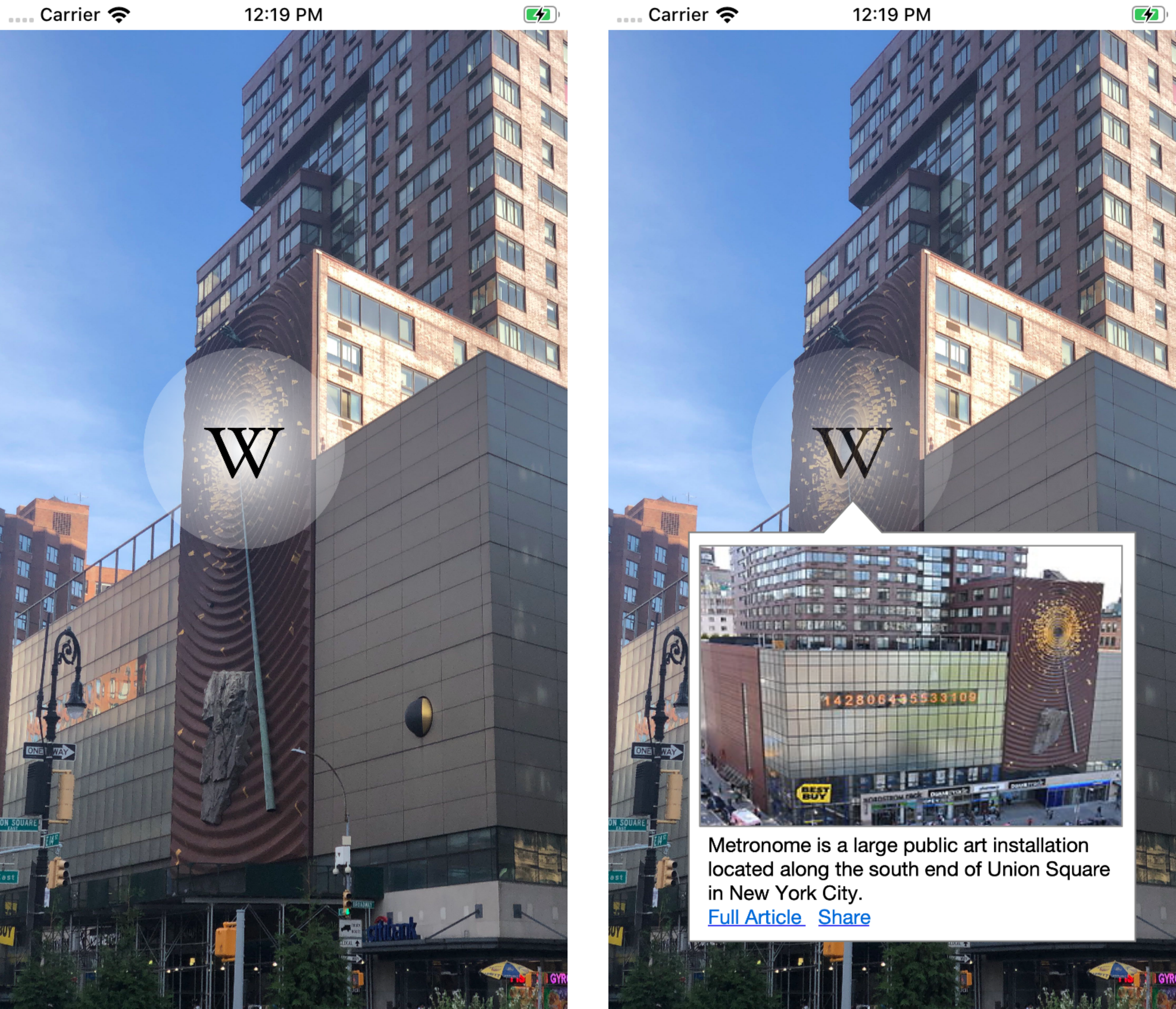}
\caption{Screenshots corresponding to the user facing mobile application. Upon recognition of a site, an active icon appears. The user can click it to access information related to the detected site.}
\label{appshots}
\end{figure}

\subsection{Overview}
With our goal being the real time recognition of notable sites in
urban environments, we next describe the development of a mobile
system that is trained on heterogeneous sources of crowd-sourced
imagery to perform the task.
The use case we are interested to support is that of a person moving in a city willing to serendipitously discover notable sites of historic, political, cultural or other interest. Wikipedia is the most well known source of knowledge describing such sites and consitutes a primary data source in the present paper. Wikipedia, in addition to offering a concrete set of notable geo-tagged items that we use to bootstrap our system, it offers relevant content that we can project to a user's device immediately after recognition of a site has taken place. 
In Figure~\ref{wikimapnyc} we show a map of geo-tagged Wikipedia articles in the area overlooking Greenwich Village in New York City. A diverse set of landmark and place types is covered with numerous points concentrated at relatively small urban areas making the recognition task challenging. The height of each polygon is proportional to the number of views the corresponding Wikipedia article had on the web.\footnote{\url{https://tools.wmflabs.org/pageviews/}}.

A primary requirement for the development of the system described above is the availability of high quality image training data. That is, data that sufficiently describes in quantity and quality terms a given site, so that it can effectively be discriminated with respect to other similar sites nearby. 
For the purposes of training we employ two source of image data, Google Images~\cite{googleImages} and Flickr~\cite{flickr}. For a large fraction of sites described on Wikipedia there is lack of sufficient image data from the Wikimedia Commons~\cite{wikicommons} collection to train effectively deep learning models for the task of interest. 
We provide a deeper look into the specification of the training data in the next paragraph, whereas we discuss a de-noising method that automatically ignores irrelevant imagery during training in Section~\ref{onlineexperiments}.

We use a convolutional neural network architecture to conduct site recognition. One limitation of deep neural network is that the set of output classes cannot be changed without retraining the network. It will also be difficult to train a single model that classify all sites in a large area. 
To make our system scalable to large areas, we separate the problem into small regions, and we train a model within each of them. Those trained models are hosted on a server and are distributed to a user's device as they enter a new region. Given the mobile setting of the deployment considered here, an equally important requirement is optimal resource utilization given typical constraints on mobile, in terms of memory, computational capacity, and internet bandwidth. Consequently, a standard deep neural network architecture trained on a large image collection would not be appropriate in this scenario. We consider therefore, a lightweight architecture, \textit{Squeezenet}~\cite{iandola2016squeezenet}, that requires a small number of training parameters while maintaining at the same time performance that matches standard state-of-the-art architectures for image classification. We review the exact details of our approach in the next paragraphs.

In Figure~\ref{appshots} we present two key instances of the prototype mobile application we develop. We assume that the app has already loaded the model from our server. First, as the user points their camera towards a target site the loaded deep learning module in the application recognises it and an active icon appears in the screen informing the user that something of potential interest has been detected. The user can then click on the icon and a short description of the site appears as retrieved from the corresponding Wikipedia article page on the Web. 


\begin{figure*}[t]
\centering
\includegraphics[scale=0.35] {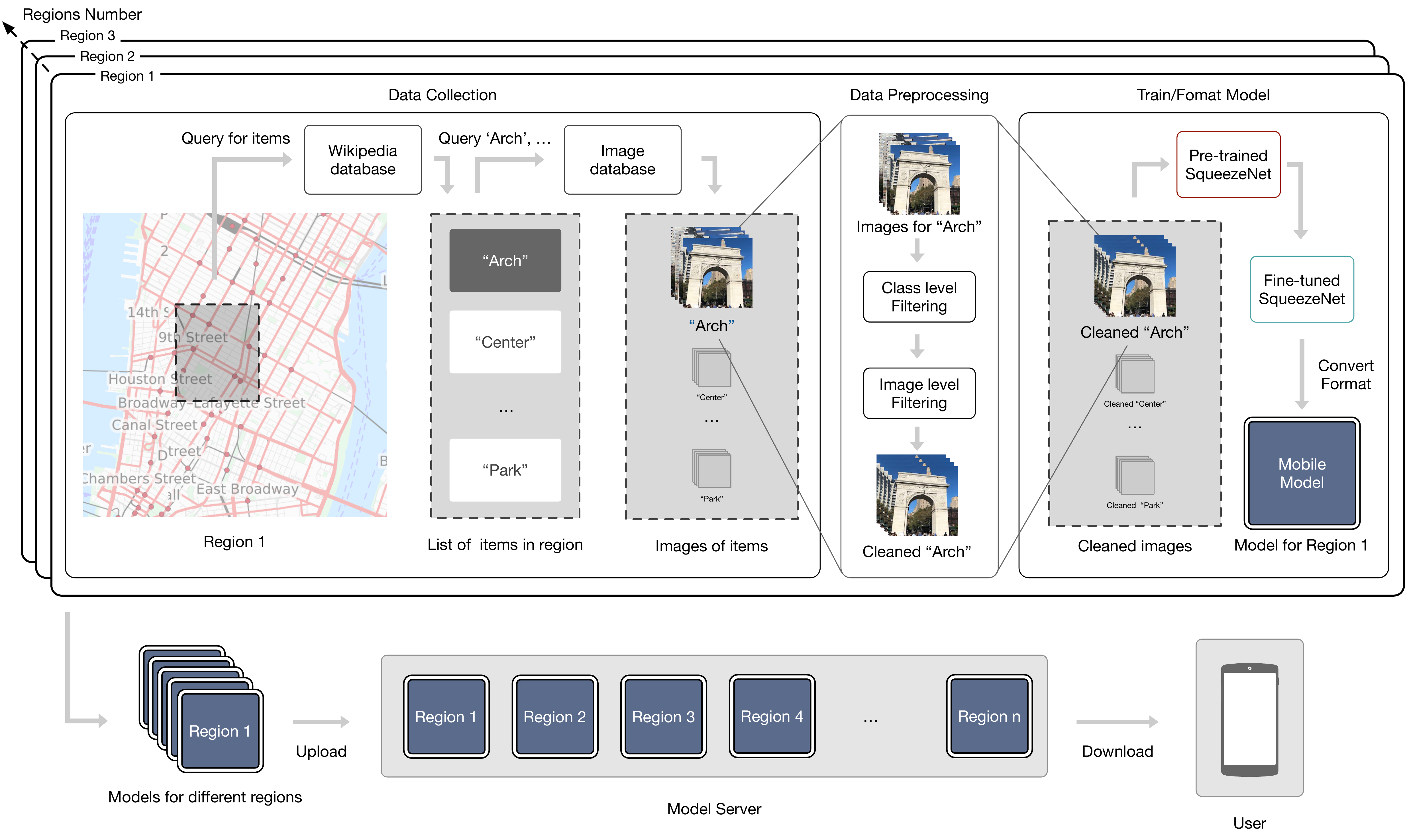}
\caption{High-level view of the end-to-end mobile site recognition system pipeline. For each site in the Wikipedia article collection of geo-tagged articles in New York, we collect relevant image data for training. During data processing a filter is applied to remove any irrelevant imagery. Finally a region specific model is trained and transmitted to the users device.}
\label{system}
\end{figure*}

\subsection{Datasets}
As noted, our work is founded on three sources of data: Wikipedia, Flickr and Google Images. We describe them formally here. In total, there are approximately $11,000$ geo-tagged Wikipedia items in the greater New York area (five boroughs) that we focus our experimentation in this work. We collect those from the Wikipedia dump portal here~\cite{wikiDumps} focusing on the English Wikipedia site. Each item in the collection is described by a title (e.g. Union Square), its  geographic coordinates, the number of views it had on Wikipedia until the last month of 2018 and its category (e.g. building, statue, church, museum etc.). Our goal then becomes to collect representative image data that describe each of those items in visual terms. We have considered two sources. First, Google Images, where we query the Google Image Search API with the title of each article and collect the first $80$ images returned~\cite{googleImages}. Our second source has been the image hosting service Flickr where we make use of the  corresponding dataset available in~\cite{flickrdata} (The Yahoo! Webscope Yahoo Flickr Creative Commons 100M, YFCC-100M ). Flickr has been a popular service for photographers touring the world with the majority of the images there being geo-tagged. We then match using site title information, the Wikipedia items with the Flickr photos dataset to extract a collection of $80$ photos for each site. In the case where a site features less than $80$ images, if it is for instance a less popular item, we simply use the number of images available in each source.

\subsection{System Specification}
In Figure~\ref{system}, we present an overview of our system's architecture. It comprises of five interconnected parts: region selection, data collection and filtering, model training, model transmission on mobile and the mobile application itself. 
Given a large area in Manhattan, we split it into small regions which are overlapping so a user experiences a soft transition as they move from one area to another. 
These region boundaries are used to query Wikipedia database for a set of sites in range. We then query image database to collect a set of images for each sites.
The crowdsourced image data collection is passed through a data filter to remove irrelevant data. The refined training set is used to build a region specific deep learning module that recognise sites which lie in that region. Finally, those regional models are transmitted from the server to the mobile application accessed by the user.  

More formally, for each region $r$, we train a different deep learning model $M_{r}$ on imagery $I$ for all sites $S_{r}$ that belong to that region $r$. We note the two sources of training data for our input imagery as $I_g$ for Google Images and $I_f$ for Flickr. The decision to build a separate model for each geographic region $r$ makes the site recognition task easier as there are fewer items for the classifier to discriminate across. The trade-off introduced by this decision is that a new model has to be off-loaded to the user's device each time they enter a different region. We take this into account by incorporating the Squeezenet model~\cite{iandola2016squeezenet}. The Squeezenet architecture achieves a lower number of parameters by reducing the size of convolution filters and attains high accuracy through downsampling late in the network to induce larger activation maps for convolution layers. One of the advantages when using a lightweight architecture, is the low bandwidth requirements that allow for quick \textit{over-the-air} updates which are desirable in a mobile setting due to well known constraints in user connectivity. Squeezenet achieves AlexNet-level accuracy with 50-fold fewer parameters and model compression techniques that allow for models that require less than 1MB in memory resources. During experimentation we have also considered deep neural network distillation techniques described by the authors in~\cite{hinton2015distilling}, though we have reduced that choice as it yielded similar accuracy scores and involved a more elaborate and complex training methodology in computational terms.

\paragraph{\textbf{Training methodology}}
Given an area $r$, we consider a collection $I_s$ of images for each site $s \in r$. The resulting training set for the area is then $I_r$, where $I_r = \{I_{s1}, I_{s2} \dots, I_{sn} \}$ considering all $n$ sites that geographically fall in the region, is then used
to train an area specific Squeezenet model. Using Cross Entropy Loss as our optimisation criterion~\cite{zhang2018generalized}, for a given training input, we seek to minimize the following loss function:
\begin{equation}
L =  - \sum_{x \in \mathcal{X}} p(x) \cdot log(q(x))
\end{equation}
where $x$ is a class in target space $\mathcal{X}$, and $p$ is the ground truth probability distribution across the classes, set equal to $1.0$ for the class corresponding to the site of interest and $0.0$ everywhere else. $q$ is the probability distribution generated by the model through a forward pass of the input image (softmax). We minimise $L$ throughout the training set and applying stochastic gradient descent and backpropagation. The model's parameters are then altered step-by-step with the goal being a higher prediction performance in the test set. A rectified linear unit (ReLU) activation function is applied after each convolution layer and prior to pooling. Each regional model has been pre-trained on the ImageNet dataset~\cite{deng2009imagenet}, prior to training on the image collection $I_r$ that is used to train a model tailored for a specific region $r$. 

\paragraph{\textbf{Model migration on Mobile}}
The training process has been implemented using the PyTorch machine learning library~\cite{paszke2017pytorch} with each regional model being trained on a cluster of GPUs for a duration that is in the order of tens of minutes. We then incorporate the trained model on the iOS and Android application platforms following two steps: first we export the model generated by PyTorch to an ONNX~\cite{onnx} format and subsequently we convert it from ONNX to a Core ML and Tensorflow Lite models~\footnote{\url{https://developer.apple.com/documentation/coreml}}\footnote{\url{https://www.tensorflow.org/lite/models/image_classification/android}} compatible with the iOS and Android platforms respectively. ONNX is an open format to represent deep learning models and allows for easy transfer across specific platforms that may adhere to different programmatic configurations. Having a version of the model on a mobile device, the system is able to perform the desired task which is to classify an image for a given site observed by the user.

\section{Online Evaluation}
\label{onlineexperiments}
In this section, we evaluate our system considering an online scenario that focuses on training and testing over the collections of crowdsourced data collected from Google Images and Flickr. We present a simple technique that enables filtering irrelevant data items from the training set so as to boost performance in the site recognition task.
The metric we employ is classification accuracy at top-$1$ which is a \textit{natural} metric to consider given the use case under consideration. By accuracy at top-$1$ we mean the fraction of times the classifier has correctly elected the ground truth label, corresponding to the site in question, at the top of the output prediction list. We remind that the classification output is based on softmax and so for every class we obtain a probability estimate. 
We consider eight areas of size $1000$ by $1000$ $m^{2}$ across central Manhattan.
Our results are obtained through averaging (mean) across the regions. Within each region we obtain a score by performing a Monte-Carlo cross validation~\cite{xu2001monte} using an $80$/$20$ ratio. We have used a learning rate of $\eta = 0.001$.

\subsection{Utilizing Crowdsourced Imagery}
\begin{figure}[t]
\centering
\includegraphics[scale=0.5] {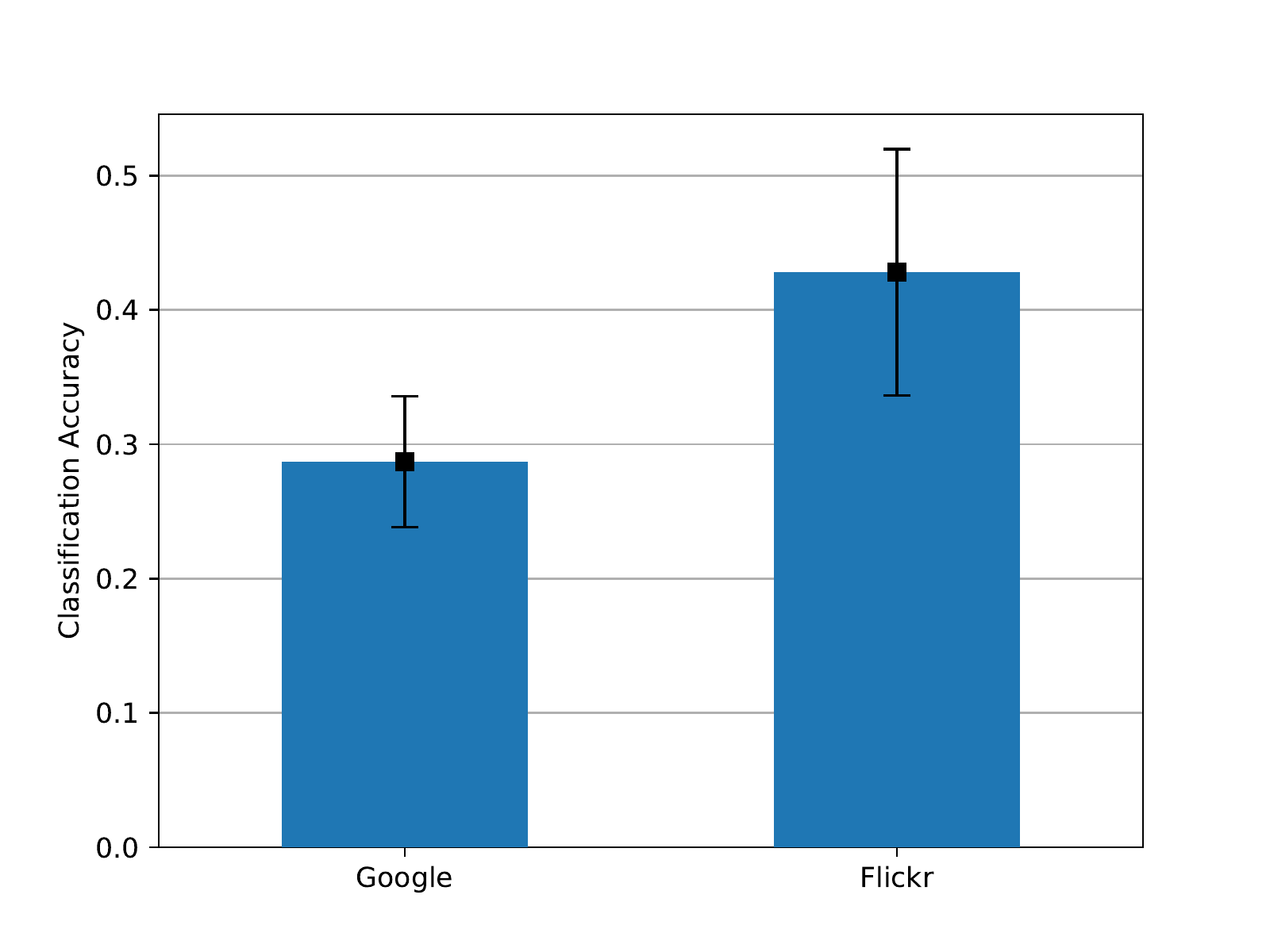}
\caption{Comparison of classification accuracy of two different data sources of training data on the site recognition task.}
\label{dsource}
\end{figure}

\paragraph{\textbf{Evaluating Data Sources}}
Given an area of size $1000$ by $1000$ $m^{2}$ we train two separate models $M_r$ and $M_r\prime$, using the set of images $I_f$ sourced from Flickr and the set $I_g$ sourced from Google. As demonstrated in Figure~\ref{dsource}, Flickr allows for better recognition performance by a significant margin of $\approx 15\%$, which corresponds to an overall improvement of almost $50\%$ compared to a model using Google Images as source. Through manual inspection we have observed that Google Image search yields results that very frequently do not correspond to outdoor views of a site. Instead a mix of image types is returned including those that provide indoor views to a venue or cases of images that are associated with generic search results, but do not necessarily provide a direct depiction of the site in question. On the other hand Flickr, being a service that has been naturally used by tourists and photographers who navigate, explore and record the city provides a more homogeneous and better curated base that we use to build our models. Even so, images obtained by Flickr also feature limitations. To alleviate the effect of noisy or irrelevant inputs we present a method that generalises to any source of crowdsourced data next.
~

\paragraph{\textbf{Image Data Purification}}
A number of geo-tagged items in the Wikipedia dataset can be of low popularity or can be abstract. This family of items can correspond to events or places that are not  visible to a navigating mobile user. As an example, let us consider the \textit{International Juridical Association}~\footnote{\url{https://en.wikipedia.org/wiki/International_Juridical_Association}} which is a professional society of lawyers established in New York City between 1931 and 1942. Online images about this Wikidata entity correspond to notable people associated with the organisation or events it hosted. An image search on this item can fetch numerous irrelevant images that results in a chaotic image dataset unrepresentative of any item present in the physical world. To make best use of crowdsourced imagery for the site recognition task, we would therefore need an automated method to filter out irrelevant image training items that could confuse a classifier.  

\begin{figure}
\begin{subfigure}{0.4\textwidth}
	\centering
	\includegraphics[scale=0.3] {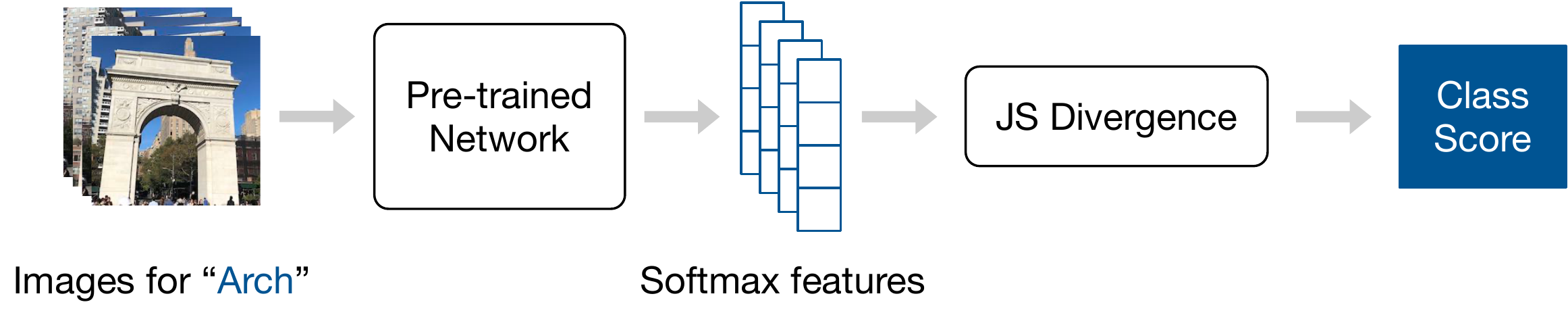}
	\caption{Class Cohesion Assessment: Given the set of training items in a class we use a Jensen-Shannon Divergence on the feature set of each item to assess how cohesive is a training set for a site.}
	\label{JSD}
\end{subfigure}
\begin{subfigure}{0.4\textwidth}
	\centering
	\includegraphics[scale=0.3] {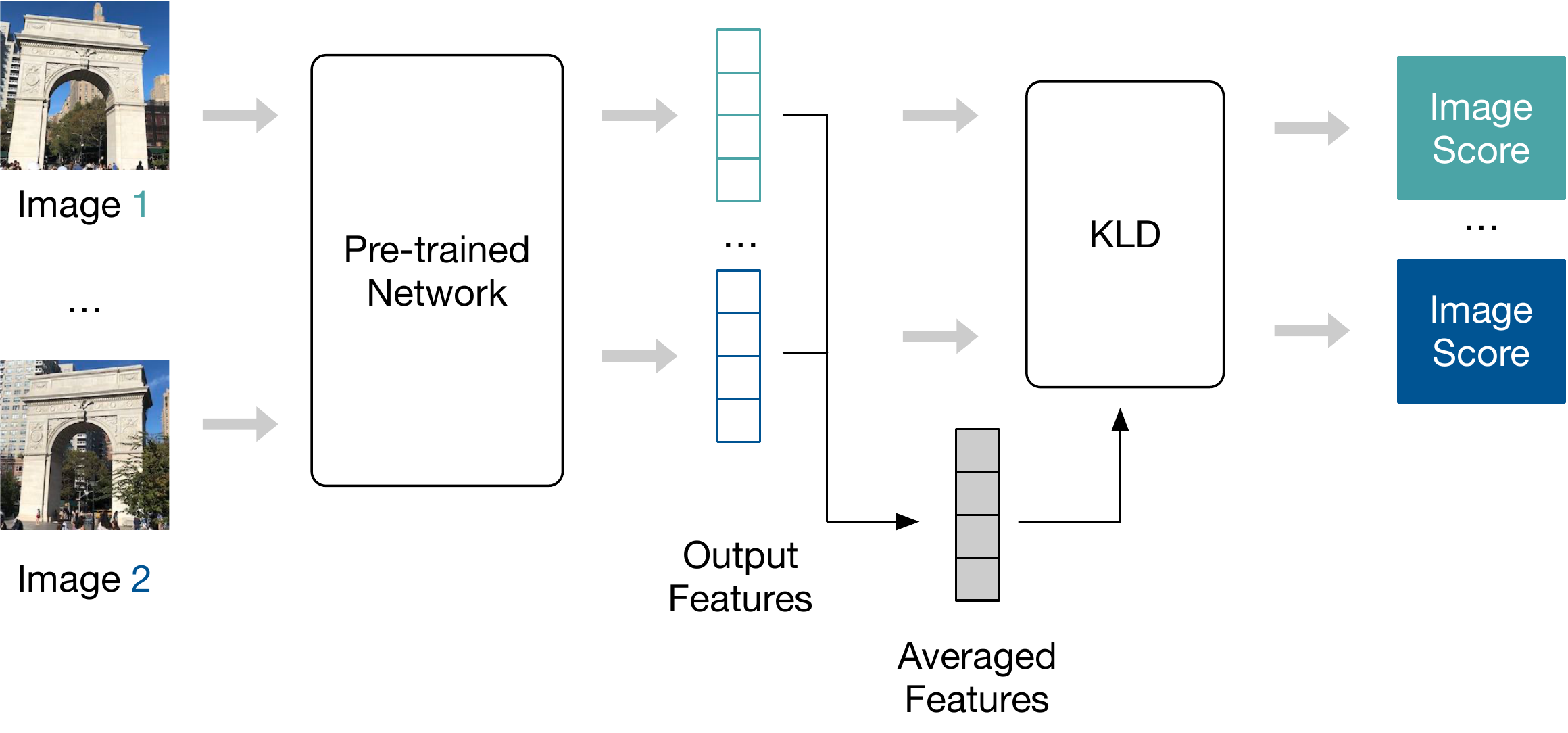}
	\caption{De-noising a class: We measure the K-L Divergence between a given training item and the centroid of the training set. Outlier items with large dirgence values are removed so that a more cohesive image-based representation of a class is attained.}
	\label{KLD}
\end{subfigure}
\caption{Training set de-noising in two steps.} 
\label{fig:purification}
\end{figure}

To do so, we devise an unsupervised filtering method with the goal being to \textit{purify} a dataset of images describing a site improving this way recognition performance.
First, given the set of images $I_s$ available for a given site $s$ we would like to perform an overall assessment regarding the appropriateness of including a site as a target for recognition in our app. If a significant majority of images associated to a Wikipedia article are irrelevant to one another, then probably the article does not correspond to a site visually recognised by users and can be removed from the set of sites considered for recognition. 
An important limitation of this assumption is that at times there are sites such us restaurants or event spaces which feature a homogeneous set of images, yet those are not associated with outdoor views of navigating users.
We describe the methodology we have followed schematically in Figure~\ref{fig:purification}. Initially, for every collected image item $i$ describing a site, we obtain a set of features $\mathbb{P}_i$, which is equal to the softmax output of Squeezenet pre-trained on ImageNet data. Taking into account then all $n$ outputs corresponding to the training image items for a site we end up with a series of $1000$-dimensional softmax vectors where each dimension corresponds to one of ImageNet's $1000$ categories. 

\begin{figure}[t]
\centering
\includegraphics[scale=0.5] {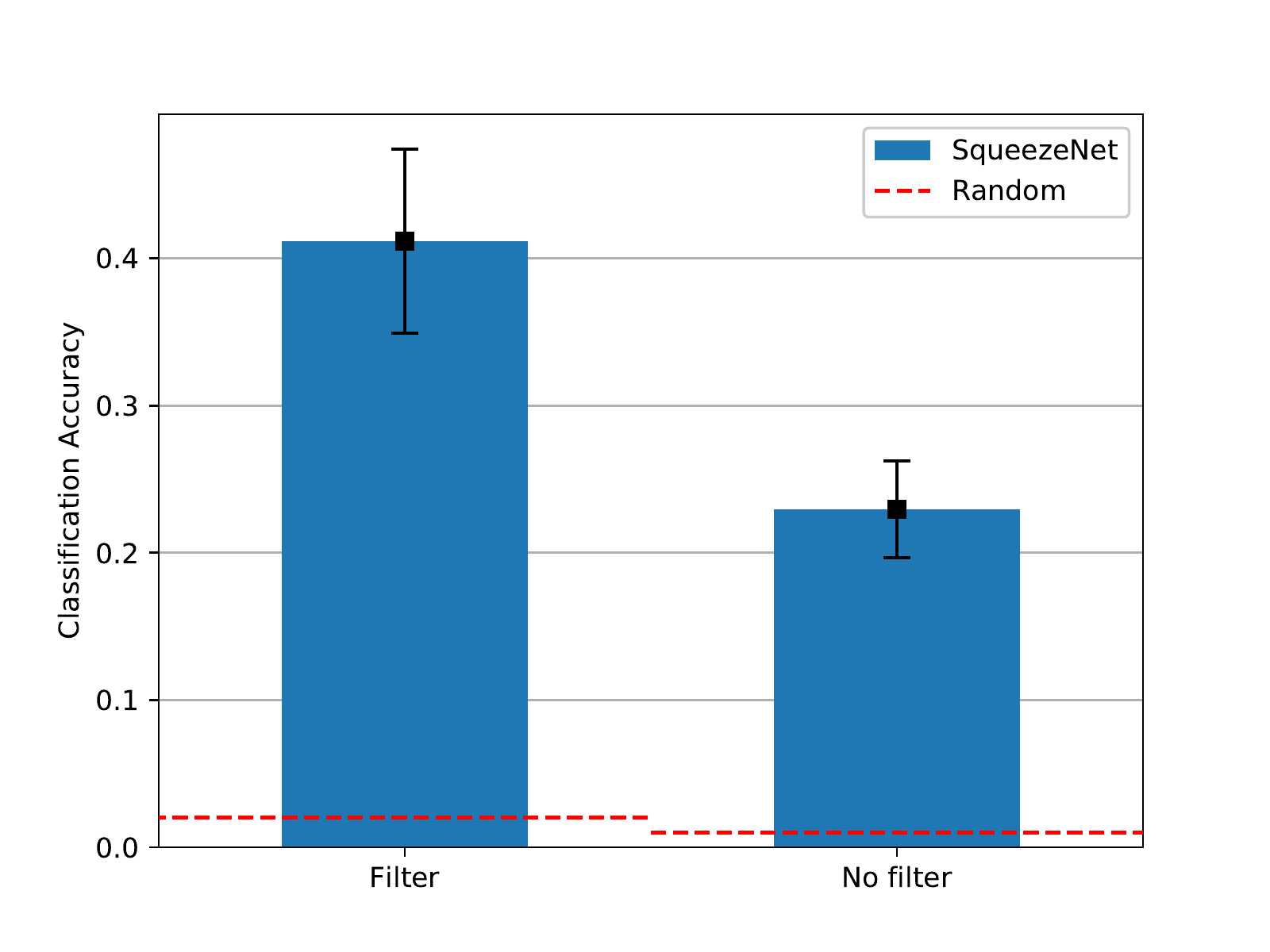}
\caption{Prediction accuracy by applying Jensen-Shannon Divergence based filtering versus the case where the filter is ignored during training.}
\label{filter}
\end{figure}

Our aim then becomes to assess the degree of \textit{cohesiveness} of this distribution in the set of training images. If the image set is too chaotic we remove the site from consideration. If however the degree of cohesiveness is high enough, we assume that there is representative training data for the item in question and we include it.  Since the softmax function outputs a probability distribution, we choose the multivariate form of Jensen-Shannon Divergence~\cite{lin1991divergence} as our measure of dispersion for a given  set of images corresponding to a site (class). Let $\mathbb{P}_1 \cdots \mathbb{P}_n$ represent the features of $n$ images in that class, and let $\pi_1 \cdots \pi_n$ represent the coefficient for each feature. The Jensen-Shannon Divergence for this image class can be defined as :
\begin{equation}
\text{JSD}_{\pi_1, \pi_2 \cdots \pi_n}(\mathbb{P}_1, \mathbb{P}_2, \cdots \mathbb{P}_n) = 
\text{H}(\sum_{i = 1}^n \pi_i\mathbb{P}_i) - \sum_{i = 1}^n \pi_i\text{H}(\mathbb{P}_i)
\end{equation}
\noindent where H is the Shannon entropy.
~
We treat each class as being of equal importance (weighting) and therefore we set each $\pi$ to $\frac{1}{n}$. As a result Equation 2 is simplified to:
\begin{equation}
\text{JSD}(\mathbb{P}_1, \mathbb{P}_2, \cdots \mathbb{P}_n) = 
\text{H}(\frac{1}{n}\sum_{i = 1}^n \mathbb{P}_i) -\frac{1}{n} \sum_{i = 1}^n \text{H}(\mathbb{P}_i)
\end{equation}
~
\noindent We then filter out all image classes which have JSD score higher than $2$. 
~

We then aim to de-noise the set of crowd-sourced images
within each of the remaining image classes. Our aim is to keep relevant images for a class while discarding outliers. In this problem setting, we define relevant type as the type of majorities. We represent the relevant feature as $\mathbb{P}_m$. 
The relevant type feature is calculated by averaging the features of all images in the class:
\begin{displaymath}
\mathbb{P}_m = \frac{1}{n}\sum_{i = 1}^n \mathbb{P}_i
\end{displaymath}
The distance between each image feature $\mathbb{P}_i$ and the relevant type feature $\mathbb{P}_m$ can be defined as the forward Kullback-Leibler Divergence between the two distributions:~
\begin{equation}
\text{D}_{\textbf{KL}}(\mathbb{P}_i || \mathbb{P}_m) = 
- \sum_{x \in \mathcal{X}} \mathbb{P}_i(x) \frac{\mathbb{P}_m(x)}{\mathbb{P}_i(x)}
\end{equation}
\noindent Images with KLD score greater than 2 are removed from the data. 
While the process of calculating JSD and KLD adds extra computational cost during the training phase, its employment has experimentally yielded superior results. In Figure~\ref{filter} we plot the prediction accuracy at top-$1$ for our model for two scenarios: with and without filtering. 
As can be observed incorporating the filter during training improves the baseline performance by almost $90\%$. The error bars show variation across the $8$ areas we have considered in our evaluation. On average the algorithm filtered out $32\%$ of the classes and $58\%$ of the total set of images.

\begin{figure}[t]
\centering
\includegraphics[scale=0.5] {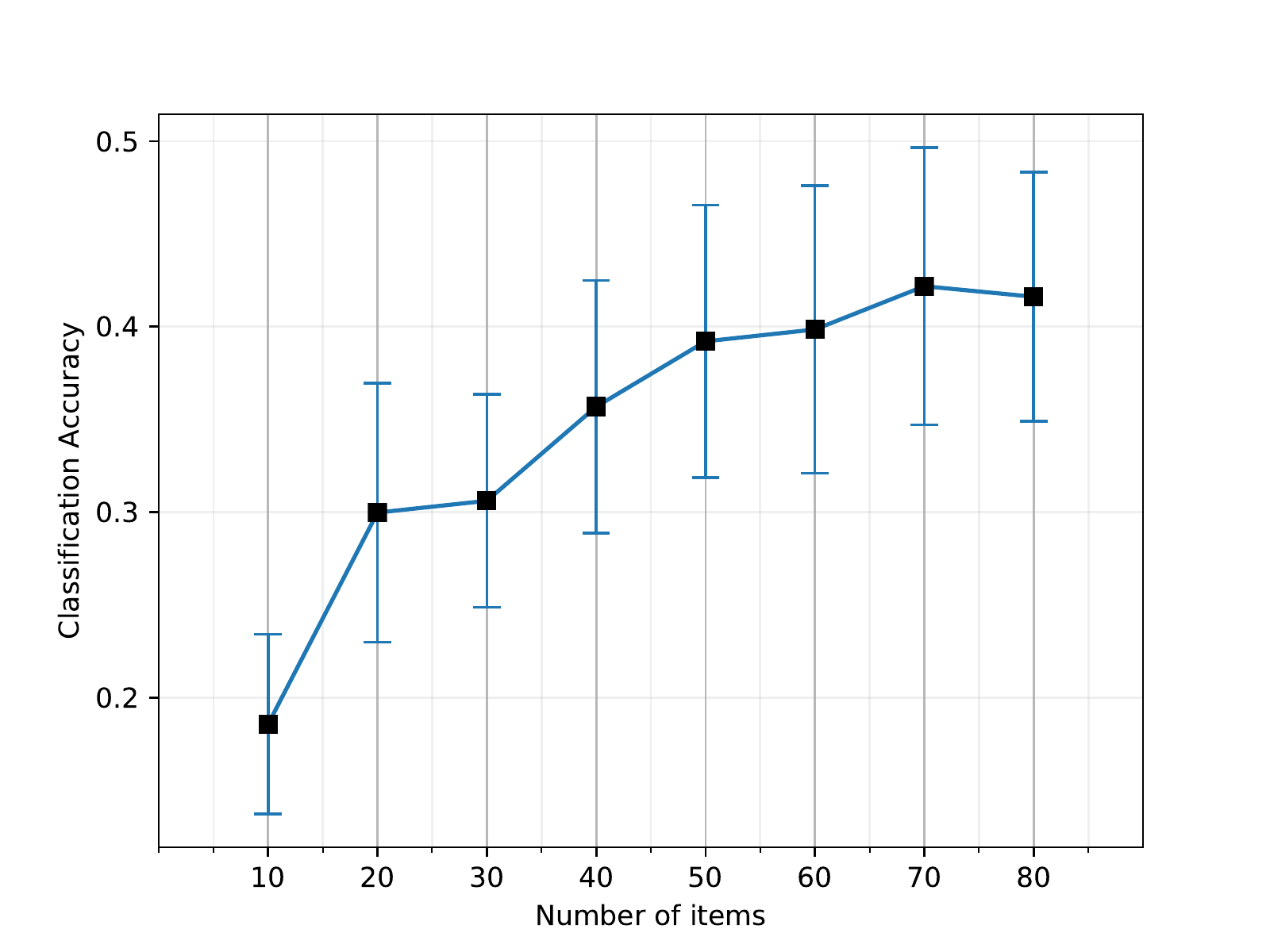}
\caption{Classification accuracy at top-1 when varying the number of image items per site used for training.}
\label{itemsperclass}
\end{figure}

The number of images in the training set available for each site is of high importance however. We demonstrate this by controlling for the number of images $m$ available when training a site. If a site has more than $m$ images, we randomly sample and select $m$ without replacement. 
In Figure~\ref{itemsperclass}, we observe that a plateau in performance in reached only when the number of training images per site available for training arrives at $70$threshold (error bars show variations across regions). Items with less images in our dataset will be less well represented for classification. Removal of irrelevant training items, as described above, should happen carefully. Removing arbitrarily image data comes with a significant penalty in performance as we have empirically demonstrated here.

~

\section{Mobile Context Awareness}
\label{sec:mcontext}
\begin{figure}[t]
\centering
\includegraphics[scale=0.5] {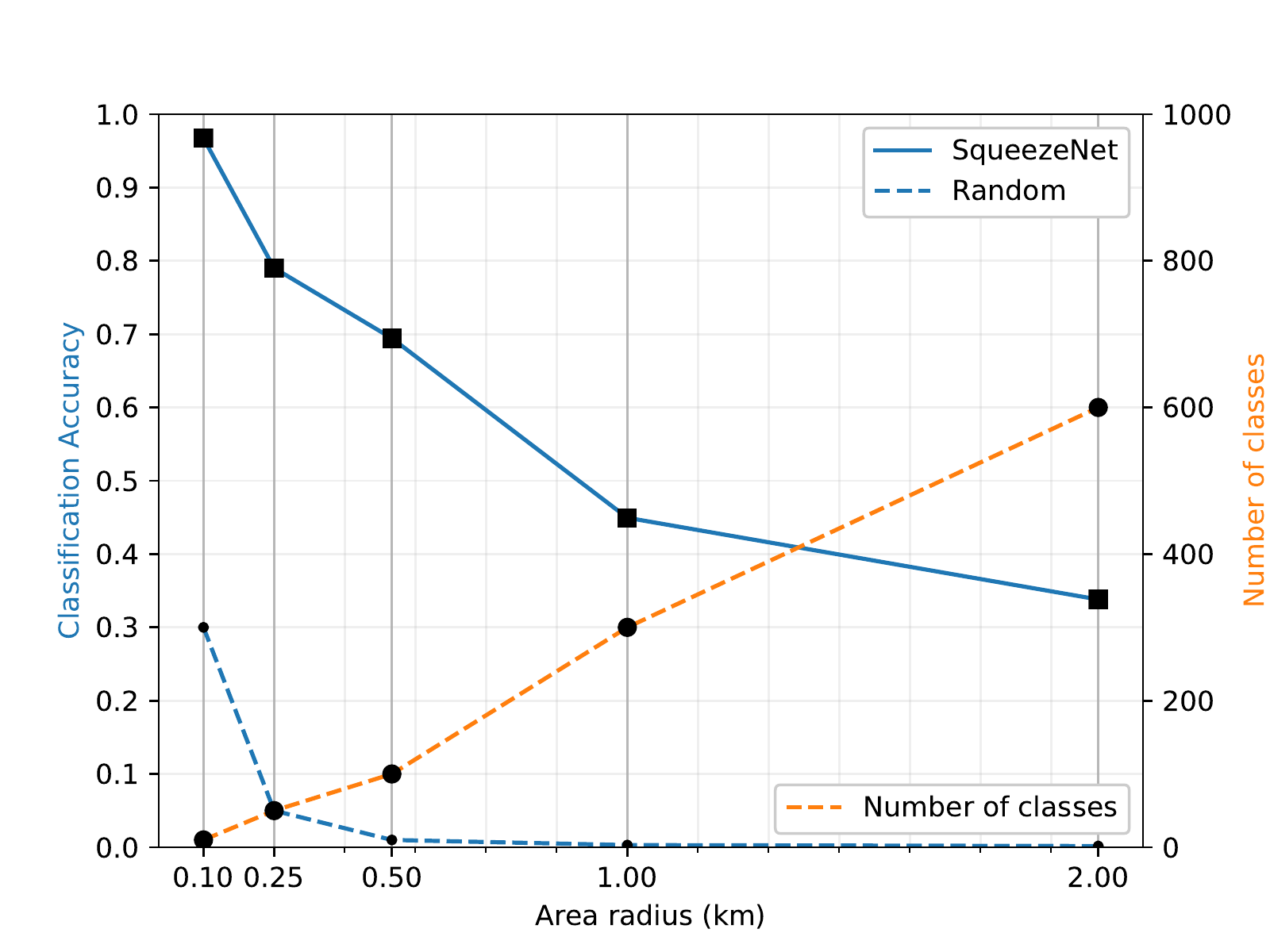}
\caption{Classification accuracy at top-1 considering different geographic area sizes.}
\label{areasize}
\end{figure}

The analysis performed so far has demonstrated that large amounts of \textit{high quality} image data data is important to the site recognition task. As we discussed, quality can be determined according to a number of dimensions, including the source of image data (e.g. Flickr), the number of training items per class or through the removal of noisy items using the data purification method we described earlier. Even so, perfectly recognising sites purely on visual terms is a very challenging task. Buildings, statues, churches and other built structures can share similar architecture, design and style features. Moreover, the fact the system is deployed in a mobile setting means that visual inputs may be prone to noise due to environmental and weather conditions or issues related to camera manipulation on the site of the user. Next, we describe the methodology we have followed towards facilitating the visual recognition task through the incorporation of mobile contextual factors. Our hypothesis is that the cooperative interplay of vision learning and mobile contextual factors can lead to superior system performance when these systems are deployed in realistic user scenarios.
\paragraph{\textbf{Determining area size for over-the-air updates on mobile.}}
One of the first consideration to make while deploying the recognition system on mobile has to do with the size of the region $r$ for building a corresponding model $M_r$ for. A too small area size, e.g. imagining the extreme scenario where we would build a different model for each site, would mean very frequent and potentially unstable model updates due to the high density of sites in certain regions. 
Coupling this with GPS signal location uncertainty can lead to inconsistent mobile update that can signifcantly harm the quality of the user experience. On the other hand, very big region sizes would translate to a large number of sites under consideration that would make site recognition harder. 
To evaluate this trade-off, we consider different geographic region sizes and perform an evaluation on prediction accuracy consiering concentring circles of increasing radius size. In Figure~\ref{areasize}, we plot classification accuracy against different geographic area sizes used to build a model $M_r$ for a given region $r$. Accuracy steadily drops as we consider larger areas with a rate of $15-20\%$ every time we double the area size. As noted, very small areas contain fewer class items and naturally the recognition task becomes easier in this setting. 
Considering the fact that accuracy levels remain relatively high ($70\%$) for a somewhat large exploration region of $1000$ by $1000$ $m^{2}$ and assuming that our users will explore the city on foot, we propose this value to fine tune the system we deploy here. However, this parameter could be fine tuned according to variations in the application scenario or when deploying to new cities of varying density. 
~
\paragraph{\textbf{Motivating the need for using mobile context in site recognition.}}
To demonstrate one of the main challenges in the site recongition task using imagery only, in Table~\ref{table1} we present the confusion matrix for top site categories during evaluation. For every site category (e.g. Museum), we calculate the probability of observing a certain category for the predicted site. In this example, a Museum  is classified to a site that is a Museaum (same category) with probability $0.67$, a Hotel with $0.11$ and a Skyscraper with $0.06$. In parenthesis we note the probability that, when the predicted category is the same, then the site recognised is the actual one (correct prediction). Sites are very likely to be confused by other sites of the same category with a prominent example of this case being Churches and Buildings. Building sites are typically widespread in an urban environment and it has been one of the categories harder to classify correctly. Churches of similar architectural style tend to also be spatially clustered in cities and hence discriminating across them can be a challenge.

\begin{figure}
\begin{subfigure}{0.23\textwidth}
\includegraphics[width=\linewidth]{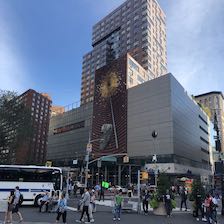}
\caption{Original Shot}
\label{metro1}
\end{subfigure}
\begin{subfigure}{0.23\textwidth}
\includegraphics[width=\linewidth]{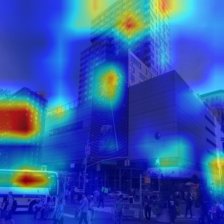}
\caption{Grad-CAM Output}
\label{metrocam1}
\end{subfigure}
\begin{subfigure}{0.23\textwidth}
\includegraphics[width=\linewidth]{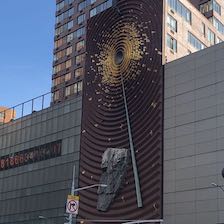}
\caption{Focused Shot}
\label{metro2}
\end{subfigure}
\begin{subfigure}{0.23\textwidth}
\includegraphics[width=\linewidth]{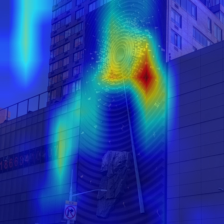}
\caption{Grad-CAM}
\label{metrocam2}
\end{subfigure}
\caption{Grad-Cam example on the Metronome Art Installation at Union Square, New York. When a user focuses on the subject of interest site classification becomes easier as nearby buildings are less likely to confuse the deep neural network.} 
\label{figgradcam}
\end{figure}

Moreover, given an input image for training or testing, a site of interest that is included in the image can be surrounded by other sites that could trigger an incorrect classification output. To demonstrate this undesirable scenario we build on the work of Selvaraju et al.~\cite{selvaraju2017grad} utilizing the Grad-CAM framework that offers visually driven explanation on the classification output of a convolutional neural network. Given an image and a class of interest, Grad-CAM produces a heatmap that highlights the features in the input image that are most important for the given class of interest. In Figure~\ref{figgradcam} we provide an example of how we have used Grad-CAM to offer interpretation given a classification output focusing on the Metronome Public Art Work at Union Square in Manhattan. The Metronome site (Fig.~\ref{metro1}) was initally classified as a building (Fig.~\ref{metrocam1}) as it is physically embedded in a built environment surrounding the site of interest. We have further observed that focusing closer to the site (Fig.~\ref{metro2}), effectively removing a portion of objects that could confuse the site recognition model, can assist the classifier towards performing a more informed guess. In the Metronome example, the site is classified correctly with visual features that are inherent to its characteristics being picked-up by the model as demonstrated in Grad-CAM output shown in Figure~\ref{metrocam2}.

\begin{table*}
\begin{tabular}{|l|l|l|l|}\hline%
\bfseries Input Category & \bfseries Output Category 1 &\bfseries Output Category 2 & \bfseries Output Category 3  
\begin{filecontents*}{confusion_probs_sorted_master.csv}
InputCategory, CategoryA, CategoryB, CategoryC
Museum,Museum 0.67 (1.0),Hotel 0.11,Skyscraper 0.06
Sculpture,Sculpture 0.62 (1.0), Museum 0.1,Nightclub 0.05
High School,High School 0.61 (0.96),Other 0.28,Post Office 0.03
Music Venue,Music Venue 0.47 (0.69),Theater 0.13,Movie Theater 0.13
Restaurant,Restaurant 0.47 (0.92),Nightclub 0.1,Other 0.08
Commercial Art Gallery,Commercial Art Gallery 0.45 (0.92),Art Museum 0.16,Venue 0.05
Building,Building 0.44 (0.92),Other 0.09,Sports Venue 0.05
College,College 0.38 (1.0),Other 0.19, Educational Institution 0.16
Movie Theater,Movie Theater 0.39 (0.96),Theater 0.1,Music Venue 0.09
Skyscraper,Skyscraper 0.36 (0.67),Other 0.12,Building 0.08
Church Building,Church Building 0.32 (0.58),Building 0.06,Other 0.06
Architectural Structure,Architectural Structure 0.47 (1.0),College 0.05,University Building 0.05
Theater,Theater 0.29 (0.79),Movie Theater 0.09,Other 0.07
Hotel,Hotel 0.23 (0.95), Building 0.12,Theater 0.12

\end{filecontents*}
\csvreader[head to column names]{confusion_probs_sorted_master.csv}{}%
{\\\InputCategory & \CategoryA & \CategoryB & \CategoryC }
\\\hline
\end{tabular}
\caption{Prediction Output Matrix of Site Categories: Given an input site category for classification we note the top-3 most likely categories output by the model. In all cases the most likely output category is the category of the input site. We note in parentheses the probability that a prediction is correct (when the output category is different the prediction is wrong).}
\label{table1}
\end{table*}

\begin{figure}[t]
\centering
\includegraphics[scale=0.45] {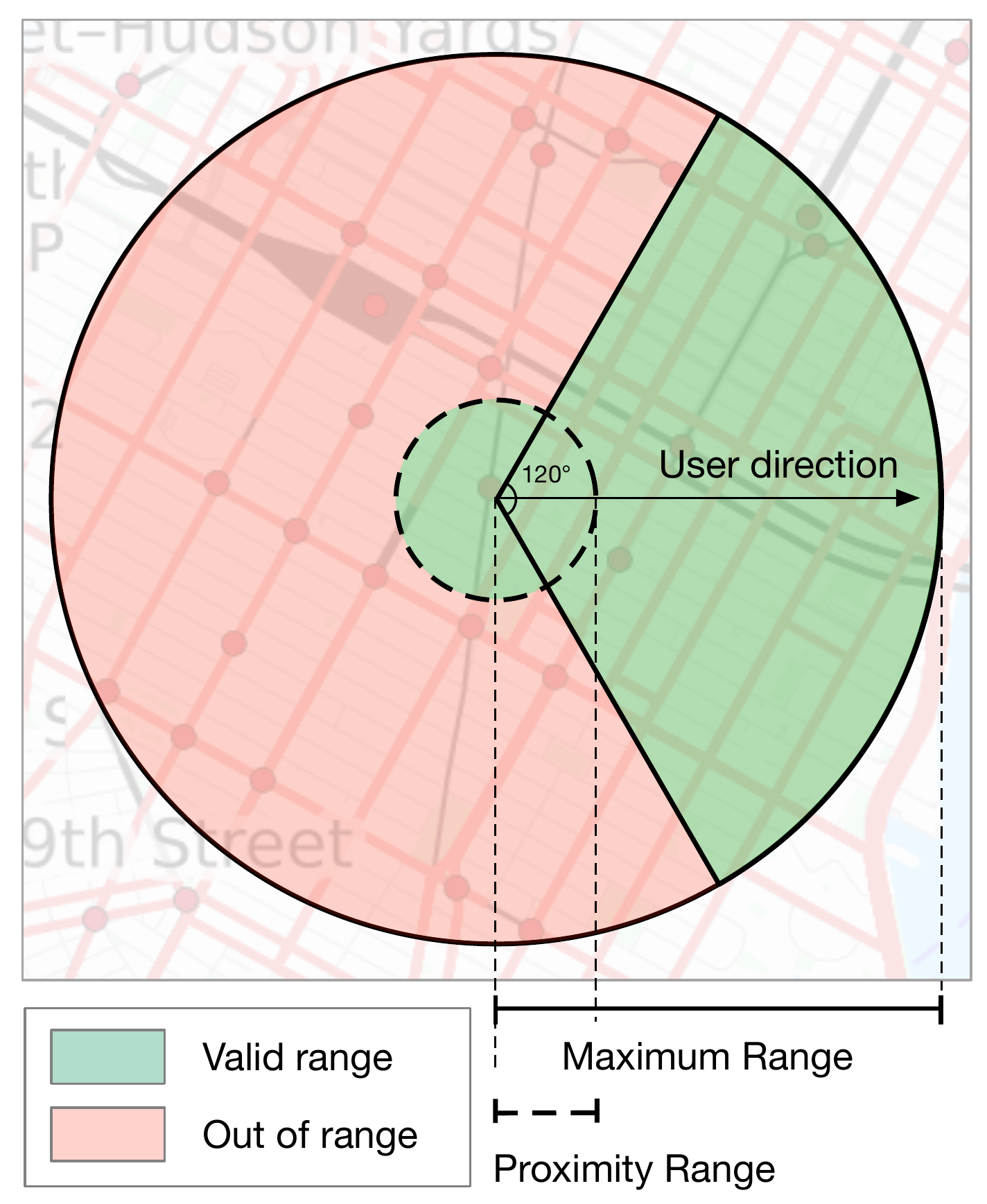}
\caption{Exploiting geo-location and user orientation to improve classification accuracy. We highlight in green the area of considered sites around the user. Out of range sites are excluded from being candidate items during site recognition.}
\label{mobilecontext}
\end{figure}

\paragraph{\textbf{Mobile Context Scheme}}
To alleviate the challenges of performing the site recongition task in the wild, we propose the incorporation of additional sources of information that capture the context of a mobile user. The scheme we are putting forward is conceptually separated in three key aspects each of which captures a different dimension of mobile user behaviour. Those are:
\begin{itemize}
\item \textbf{User geo-location:} Using the geographic coordinates of the user as provided by their phone's GPS sensor, we can filter out from the site recognition task all sites that are at a large geographic distance from the user's location. While we have effectively incorporate location information implicitly through off-loading a model to the user's device according to geography, we take an extra step to further impose geographic constraints on the task by considering sites that are within $200$ meters from the user location. The threshold comfortably accommodates GPS errors of modern mobile devices and at the same maintains a generous degree of freedom to allow the user interact with sites that are not in very close vicinity but can appear relevant (e.g. a large landmark visible from a distance).

\item \textbf{User orientation:} When a user points their phone's camera towards a site of potential interest they additionally provide directional information that can be picked-up by the phone's compass. We can therefore remove sites that are not directly visible by the user's camera through the introduction of a site validity zone based on orientation. Considering a $360$-degree angular space of orientation we allow for the consideration of sites in the classification task that lie within $60$ degrees plus or minus from the users bearing direction. This allows for an overall $120$-degree coverage which sufficiently accounts for compass sensor errors and yet captures a large portion of the observer's field of view. 

\item \textbf{User attention:} Finally, we incorporate users attention patterns to focus the classification task
by incorporating the feedback provided by the Grad-CAM visual explanation system in our mobile app. At first instance, a visual symbolic clue (Wikipedia icon as shown in Figure~\ref{appshots}) is provided to the user superimposed over the area of the image where the classifier detects a potential object of interest. The user can then tap to view related information on the recognised site. Alternatively, the user can manually select the image area of interest with a simple polygon design interface accessible through the app. A subsequent attempt for classification will be performed then by the app focusing on the user selected area with the icon being repositioned to point the user to the newly detected object.    
\end{itemize}

\begin{figure}[t]
\centering
\includegraphics[scale=0.5] {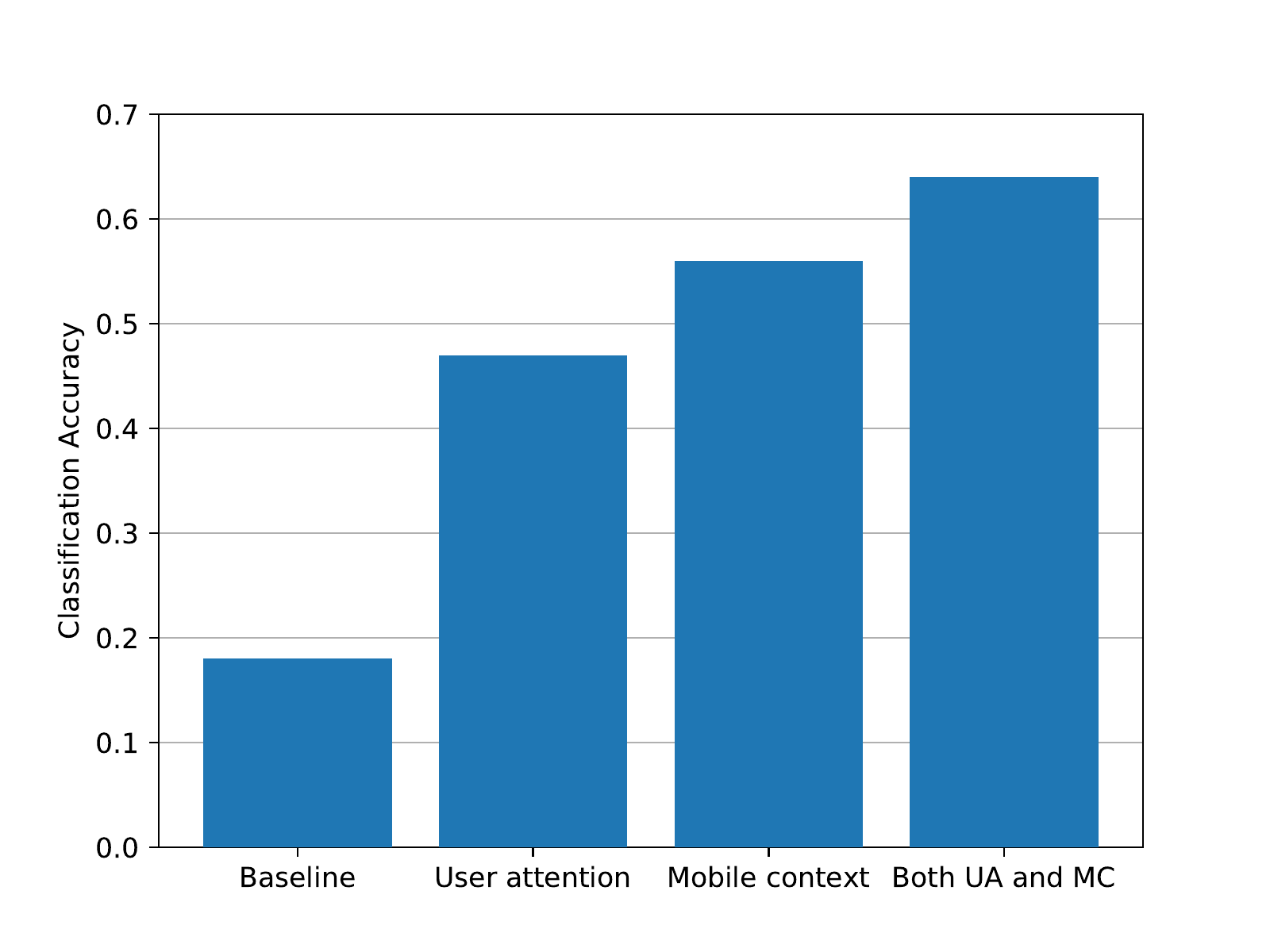}
\caption{Classification Accuracy Using Mobile Context.}
\label{mobilecontext_figure}
\end{figure}

\paragraph{\textbf{Mobile Context Evaluation}}
Our approach is schematically described in Figure~\ref{mobilecontext} with green space highlighting the geographic range around the user that contains the set of sites that considered in the classification task. We integrate this as a filtering mechanism in the mobile app simply by discarding all softmax outputs for classes (sites) that are excluded and electing the highest scoring site from the remaining candidates. The necessity for the use of mobile context to assist the visual recognition system emerged after a test deployment. We collected $100$ photos across $25$ randomly selected sites we visited in a central area of New York City and we assessed the efficacy of the classification model in this test set. Multiple photos for each site were taken approaching it from different directions and perspectives. We highlight that this test set is distinct to the Flickr dataset we used for training the classifier and our aim here has been to testbed our system over inputs captured in a real mobile context as opposed to considering an online evaluation scenario we explored in the previous Section. 
In Figure~\ref{mobilecontext_figure} we present the accuracy results obtained on this test set. Notably, the accuracy performance deteriorated significantly compared to the online scenario we presented in the previous section, dropping to a value of $18\%$. Incorporating \textit{user attention} information with the user able to focus on a subject of interest, allowed the model to rise up to $48\%$.
Using geo-location information and user information allows for the model to attain a performance of $56\%$, 
whereas using all mobile context inputs (orientation, location and user attention) information allowed for a further improvement of the accuracy in the test set to $64\%$. These results demonstrate that despite the importance of incorporating online imagery to train such systems, when deploying them in the wild new inputs induced by mobile users can introduce noise as well as inputs of image items that are sampled from an effectively distinct distribution to the one described by online datasets. Here we show that complementing the value extracted of those with mobile contextual factors can recover significantly the margin that may exist between curated online data and novel inputs introduced by mobile users. 

\section{Limitations and Vision}
\label{limitations}
Our work in the previous sections has shown that despite the ambitious task we undertook, identifying sites of interest on mobile through the use of visual clues and capturing information about mobile user context is feasible. Doing so can enable the effective exploration of Wikipedia content in an intuitive manner. This is particularly important for tourists and urban explorers who are willing to learn about interesting facts related to a city, sites and landmarks of historic and contemporary significance, as well as hidden gems in urban environments that are not typically accessible through other sources of knowledge. Interacting with urban space through new mediums can power a new generation of artificially intelligent and augmented reality applications. The capacity to perform these tasks has been enabled through progress both in the availability of computational resources in mobile devices, but also the large collections of imagery becoming available through a variety of sources. Nonetheless, we have identified key limitations to the use of online data for training deep learning modules. As we have shown, one way to resolve this challenge is to filter irrelevant images with regard to the site of interest. Even so, there will always exist less popular sites for which image data will be scarce. We believe that making these niche sites accessible through our application can be intriguing and help users become more eager for exploration and acquiring knowledge. We seek to alleviate this issue in future work, using data collected from Google Street View and similar services, though automating this process at scale is a major challenge.

There are already commercial applications such as Google Lens that are able to recognise objects and sites outdoors. Our aim in the present work is to inform the research community on the general concepts and technical directions one would follow to develop applications of this type. Moreover, we are particulary focused to helping users accessing knowledge of notable things, in line with the vision set by the Wikipedia community, as opposed to supporting use cases involving tasks such as generic item search or shopping. We will be storing all site photos collected through our application in a repository with all media donated to Wikipedia Commons~\cite{wikicommons}. This way we enable a crowdsourcing loop, where users are consumers of crowdsourced content, but at the same time they are media content donours. Aspects of gamification will also be enabled in this setting, with users being able to collect points based on the number of sites they recognise and be rewarded for contributing media content to the community.

\section{Related Work}
\label{related}
The increasing availability of computational resources, improved battery technology and communication capabilities of smartphone devices has allowed for a new generation of machine learning powered mobile services to emerge over the past decade. As a result there has been a host of works dedicated on understanding how machine and deep learning models can be deployed in this setting. Much of the focus has been on sensing user activity (e.g. walking, driving etc.) with the goal to improve physical or mental health or enabling new channels of social communication between users~\cite{miluzzo2008sensing, shoaib2015survey, harari2016using, spathis2019sequence}. Deep learning modules on mobile typically focus on tasks where these models are known to excel, such as audio sensing or visual object detection~\cite{lane2015deepear} with a challenge being the development of so called light models that have small memory footprint and quick inference responses so they can effectively be accommodated in a mobile context~\cite{lane2017squeezing, hinton2015distilling, howard2017mobilenets}. Visual search on mobile~\cite{liu2017deep} using deep learning has been another subject of study related to the present work to support applications such as that of identifying plants in nature~\cite{mohanty2016using} or object search supported by applications such as Google Lens~\cite{huang2015systems}. 

Our work relates more closely to studies on what is known as site or landmark identification. Most works in this domain have focused on predicting the \textit{tags} of specific image (e.g. statue, bridge etc.). The task we tackle here however aims at specifically identifying a unique site and provides new application opportunities associated with tourist exploration and local knowledge augmentation. That latter task has been also been studied by authors in~\cite{zheng2009tour} where a web-scale landmark recognition engine based on visual detection models is being proposed. Their approach is limited to considering only an offline evaluation scenario on a set of images corresponding to global landmarks, ignoring mobility context altogether. The authors in~\cite{tang2015improving} describe a methodology that proposes the incorporation of geographic information to improve image classification, which we draw inspiration from in this work, but focusing on notable sites of interest to urban explorers, as opposed to generic imagery content used in~\cite{tang2015improving}. The reverse task, inferring a geographic location  given an image has also been studied previously~\cite{weyand2016planet}.

\section{Conclusion}
\label{conclusion}
In this work we have developed an end-to-end mobile system that exploits deep learning on crowdsourced images to perform recognition of notable sites. We have highlighted the significance of online sources of data in bootstrapping and training such systems. We have further developed an automated filtering process that allows to better exploit image data collected by mobile users and becomes available through online photography services like Flickr. Site recognition in the wild can be challenging not only because distinguishing between sites in the physical world is inherently difficult due to item similarities and the large number of candidate items, but also due to the fact that within a user's camera visual field irrelevant objects can trigger false positive events. Using mobile contextual information such as user attention patterns, geo-location and orientation can alleviate these challenges to an extent where such applications can become usable as well as useful.

\bibliographystyle{ACM-Reference-Format}
\bibliography{biblio}

\end{document}